\documentclass[10pt,twocolumn,letterpaper]{article}

\usepackage{iccv}
\usepackage{times}
\usepackage{epsfig}
\usepackage{graphicx}
\usepackage{amsmath}
\usepackage{amssymb}
\usepackage{caption}
\usepackage{bm}
\usepackage{comment}
\usepackage[numbers,sort&compress]{natbib}
\usepackage{multirow}

\newcommand\blfootnote[1]{%
  \begingroup
  \renewcommand\thefootnote{}\footnote{#1}%
  \addtocounter{footnote}{-1}%
  \endgroup
}

\usepackage[pagebackref=true,breaklinks=true,letterpaper=true,colorlinks,bookmarks=false]{hyperref}

\iccvfinalcopy

\ificcvfinal\fi

\begin{document}

%%%%%%%%% TITLE
\title{Learning Local Recurrent Models for Human Mesh Recovery}

\author{Runze Li$^{1,2}$, Srikrishna Karanam$^{1}$, Ren Li$^{1}$, Terrence Chen$^{1}$, Bir Bhanu$^{2}$, and Ziyan Wu$^{1}$\\
$^{1}$United Imaging Intelligence, Cambridge MA, USA\\
$^{2}$University of California Riverside, Riverside CA, USA\\
{\tt \{first.last\}@uii-ai.com}
}

\twocolumn[{%
\renewcommand\twocolumn[1][]{#1}%
\maketitle
\begin{center}
    \centering
    \includegraphics[width=1.0\linewidth]{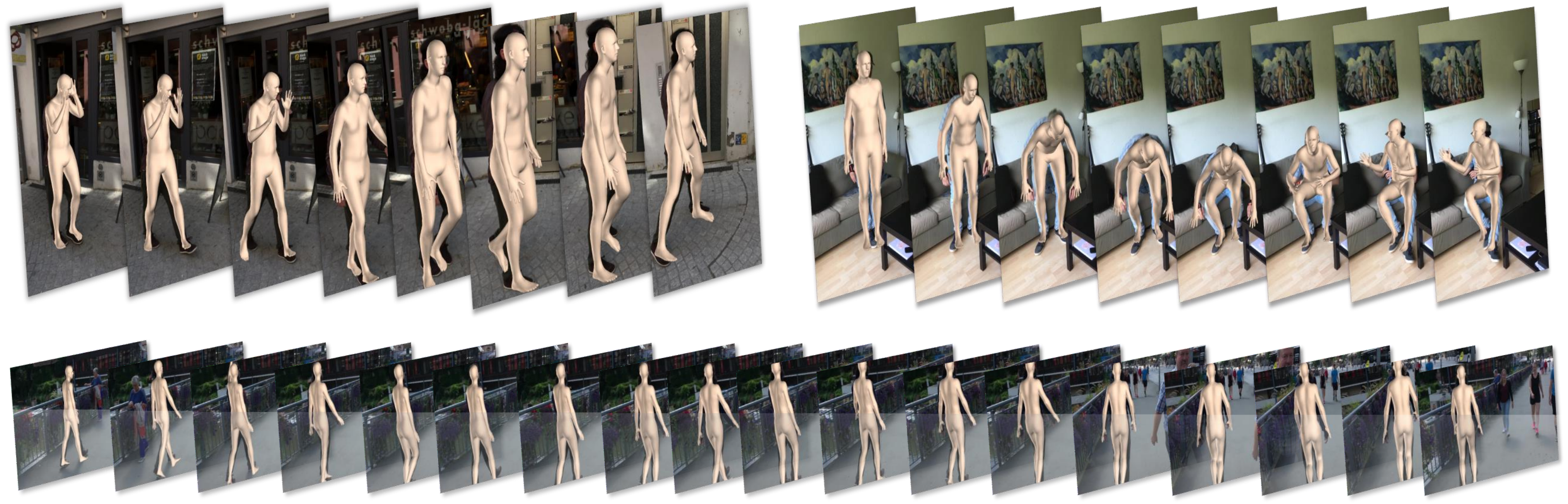}
    \captionof{figure}{We present \textbf{LMR}, a new method for video human mesh recovery. Unlike existing work, LMR captures local human part dynamics and interdependencies by learning multiple local recurrent models, resulting in notable performance improvement over the state of the art. Here, we show a few qualitative results on the 3DPW dataset.}
    \label{fig:title}
\end{center}%
}]

\ificcvfinal\thispagestyle{empty}\fi

%%%%%%%%% ABSTRACT
\begin{abstract}

We consider the problem of estimating frame-level full human body meshes given a video of a person with natural motion dynamics. While much progress in this field has been in single image-based mesh estimation, there has been a recent uptick in efforts to infer mesh dynamics from video given its role in alleviating issues such as depth ambiguity and occlusions. However, a key limitation of existing work is the assumption that all the observed motion dynamics can be modeled using one dynamical/recurrent model. While this may work well in cases with relatively simplistic dynamics, inference with in-the-wild videos presents many challenges. In particular, it is typically the case that different body parts of a person undergo different dynamics in the video, e.g., legs may move in a way that may be dynamically different from hands (e.g., a person dancing). To address these issues, we present a new method for video mesh recovery that divides the human mesh into several local parts following the standard skeletal model. We then model the dynamics of each local part with separate recurrent models, with each model conditioned appropriately based on the known kinematic structure of the human body. This results in a structure-informed local recurrent learning architecture that can be trained in an end-to-end fashion with available annotations. We conduct a variety of experiments on standard video mesh recovery benchmark datasets such as Human3.6M, MPI-INF-3DHP, and 3DPW, demonstrating the efficacy of our design of modeling local dynamics as well as establishing state-of-the-art results based on standard evaluation metrics. 
  
\end{abstract}

%%%%%%%%% BODY TEXT
\section{Introduction}
\label{sec:intro}

\begin{figure}[h!]
\centering
\includegraphics[width=1.0\linewidth]{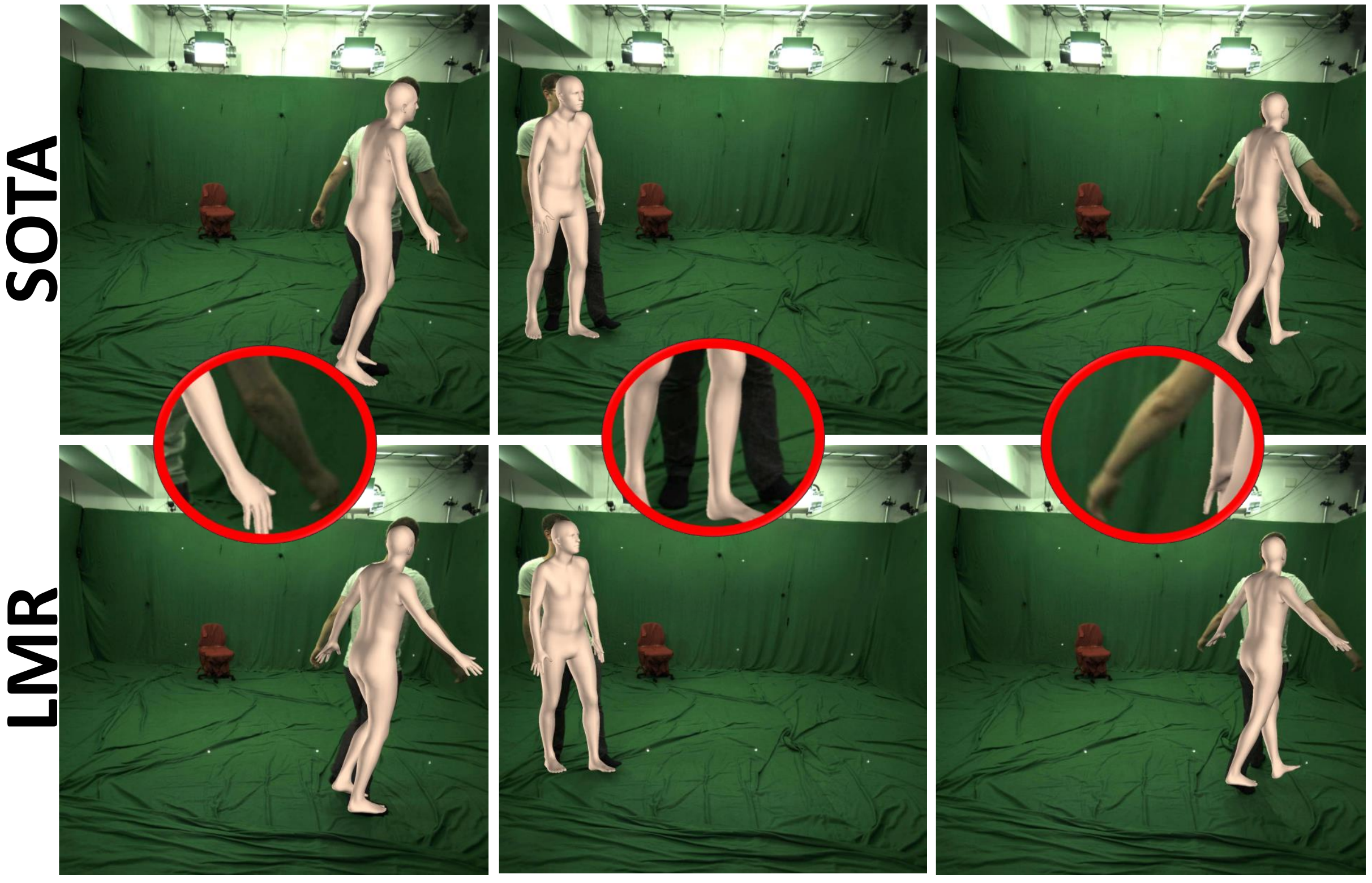}
\caption{A qualitative comparison with VIBE \cite{kocabas2019vibe}, highlighting local regions (ellipses that show zoomed-in VIBE results) where LMR gives better performance.}
\label{fig:vibecomparepage2}
\end{figure}

\blfootnote{*This work was done during the internships of Runze Li and Ren Li with United Imaging Intelligence. Corresponding author: Srikrishna Karanam.}We consider the problem of human mesh recovery in videos, i.e., fitting a parametric 3D human mesh model to each frame of the video. With many practical applications \cite{singh2017darwin,martinez2018real}, including in healthcare for COVID-19 \cite{li2007automatic,ching2014patient,karanam2020towards}, there has been much progress in this field in the last few years \cite{hmrKanazawa17,kocabas2019vibe,georgakis2020hierarchical}. In particular, most research effort has been expended in single image-based mesh estimation where one seeks to fit the human mesh model to a single image. However, such 3D model estimation from only a single 2D projection (image) is a severely under-constrained problem since multiple 3D configurations (in this case poses and shapes of the mesh model) can project to the same image. Such ambiguities can be addressed by utilizing an extra dimension that is typically associated with images- the temporal dimension leading to video data and the problem of video mesh recovery.

The currently dominant paradigm for video mesh recovery involves the \textsl{feature-temporal-regressor} architecture. A deep convolutional neural network (CNN) is used to extract frame-level image feature vectors, which are then processed by a temporal encoder to learn the motion dynamics in the video. The representation from the temporal encoder is then processed by a parameter regressor module that outputs frame-level mesh parameter vectors. While methods vary in the specific implementation details, they mostly follow this pipeline. For instance, while Kanazawa \etal \cite{humanMotionKanazawa19} implement the temporal encoder using a feed-forward fully convolutional model, Kocabas \etal \cite{kocabas2019vibe} uses a recurrent model to encode motion dynamics. However, uniformly across all these methods, the parameter regressor is implemented using a ``flat" regression architecture that takes in feature vectors as input and directly regresses all the model parameters, e.g., 85 values (pose, shape, and camera) for the popularly used skinned multi-person linear (SMPL) model \cite{loper2015smpl,hmrKanazawa17}. While this paradigm has produced impressive recent results as evidenced by the mean per-joint position errors on standard datasets (see Arnab \etal \cite{Arnab_2019_CVPR} and Kocabas \etal \cite{kocabas2019vibe} for a fairly recent benchmark), a number of issues remain unaddressed that provide us with direction and scope for further research and performance improvement. 

First, the above architectures implicitly assume that all motion dynamics can be captured using a single dynamical system (e.g., a recurrent network). While this assumption may be reasonable for fairly simplistic human motions, it is not sufficient for more complex actions. For instance, while dancing, the motion dynamics of a person vary from one part of the body to the other. As a concrete example, the legs may remain static while the hands move vigorously, and these roles may be reversed after a certain period of time (static hands and moving legs several frames later), leading to more ``locally" varying dynamics. Intuitively, this tells us that the motion of each local body part should in itself be modeled separately by a dynamical system, and that such a design should help capture this local ``part-level" dynamical information more precisely as opposed to a single dynamical system for the entire video snippet.

Next, as noted above, the \textsl{regressor} in the \textsl{feature-temporal-regressor} architecture involves computing all the parameters of the SMPL model using a direct/flat regression design without due consideration given to the interdependent nature of these parameters (i.e., SMPL joint rotations are not independent but rather conditioned on other joints of other parts such as the root \cite{loper2015smpl}). It has been noted in prior work \cite{kendall2017geometric} that such direct regression of rotation matrices, which form a predominant part of the SMPL parameter set, is challenging as is and only made further difficult due to these interdependencies in the SMPL model. In addition to direct rotation regression, the temporal module in the above \textsl{feature-temporal-regressor} also does not consider any joint and part interdependencies, i.e., modeling all motion dynamics using a single global dynamical system, thus only further exacerbating this problem. 

To address the aforementioned issues, we present a new architecture for capturing the human motion dynamics for estimating a parametric mesh model in videos. Please note that while we use the SMPL model \cite{loper2015smpl} in this work, our method can be extensible to other kinds of hierarchical parametric human meshes as well. See Figure~\ref{fig:title} for some qualitative results with our method on the 3DPW \cite{von2018recovering} dataset and Figure~\ref{fig:vibecomparepage2} for a comparison with a current state-of-the-art method. Our method, called \textsl{local recurrent models for mesh recovery} (\textbf{LMR}), comprises several design considerations. First, to capture the need for modeling locally varying dynamics as noted above, LMR defines six local recurrent models (root, head, left/right arms, left/right legs), one each to capture the dynamics of each part. As we will describe later, each ``part" here refers to a chain of several joints defined on the SMPL model. Note that such a part division is not ad hoc but grounded in the hierarchical and part-based design of the SMPL model itself, which divides the human body into the six parts above following the standard skeletal rigging procedure \cite{loper2015smpl}. Next, to model the conditional interdependence of local part dynamics, LMR first infers root part dynamics (i.e., parameters of all joints in the root part). LMR then uses these root part parameters to subsequently infer the parameters of all other parts, with the output of each part conditioned on the root output. For instance, the recurrent model responsible for producing the parameters of the left leg takes as input both frame-level feature vectors as well as frame-level root-part parameters from the root-part recurrent model.

Note the substantial differences between LMR's design and those of prior work- (a) we use multiple local recurrent models instead of one global recurrent model to capture motion dynamics, and (b) such local recurrent modeling enables LMR to explicitly capture local part dependencies. Modeling these local dependencies enables LMR to infer motion dynamics and frame-level video meshes informed by the geometry of the problem, i.e., the SMPL model, which, as noted in prior work \cite{kendall2017geometric}, is an important design consideration as we take a step towards accurate rotation parameter regression architectures. We conduct extensive experiments on a number of standard video mesh recovery benchmark datasets (Human3.6M \cite{ionescu2013human3}, MPI-INF-3DHP \cite{mehta2017monocular}, and 3DPW \cite{von2018recovering}), demonstrating the efficacy of such local dynamic modeling as well as establishing state-of-the-art performance with respect to standard evaluation metrics. 

To summarize, the key contributions of our work are:

\begin{itemize}
    \item We present LMR, the first local-dynamical-modeling approach to video mesh recovery where unlike prior work, we explicitly model the local dynamics of each body part with separate recurrent networks.
    \item Unlike prior work that regresses mesh parameters in a direct or ``flat" fashion, our local recurrent design enables LMR to explicitly consider human mesh interdependencies in parameter inference, thereby resulting in a structure-informed local recurrent architecture.
    \item We conduct extensive experiments on standard benchmark datasets and report competitive performance, establishing state-of-the-art results in many cases.
\end{itemize}

\section{Related Work}
\label{sec:related}
There is much recent work in human pose estimation, including estimating 2D keypoints \cite{newell2016stacked, cao2019openpose, zhang2020distribution}, 3D keypoints \cite{martinez2017simple,pavlakos2018ordinal,habibie2019wild, iqbal2020weakly,gpa}, and a full mesh \cite{hmrKanazawa17,pavlakos2018learning,humanMotionKanazawa19,kolotouros2019convolutional,Arnab_2019_CVPR,georgakis2020hierarchical,kocabas2019vibe}. Here, we discuss methods that are relevant to our specific problem- fitting 3D meshes to image and video data. 
\\
\indent \textbf{Single-image mesh fitting.} Most recent progress in human mesh estimation has been in fitting parametric meshes to single image inputs. In particular, following the availability of differentiable parametric models such as SMPL \cite{loper2015smpl}, there has been an explosion in interest and activity in this field. Kanazawa \etal \cite{hmrKanazawa17} presented an end-to-end trainable regression architecture for this problem that could in principle be trained with 2D-only keypoint data. Subsequently, many improved models have been proposed. Kolotourous \etal \cite{kolotouros2019convolutional} and Georgakis \etal \cite{georgakis2020hierarchical} extended this architecture to include more SMPL-structure-informed design considerations using either graph-based or parameter factorization-based approaches. There have also been attempts at SMPL-agnostic modeling of joint interdependencies, with Fang \etal \cite{fang2017learning} employing bidirectional recurrent networks and Isack \etal \cite{isack2020repose} learning priors between joints using a pre-defined joint connectivity scheme. While methods such as Georgakis \etal \cite{georgakis2020hierarchical} and Zhou \etal \cite{zhou2016deep} also take a local part-based kinematic approach, their focus is on capturing inter-joint spatial dependencies. On the other hand, LMR's focus is on capturing inter-part temporal dependencies which LMR models using separate recurrent networks.\\
\indent \textbf{Video mesh fitting.} Following the success of image-based mesh fitting methods, there has been a recent uptick in interest and published work in fitting human meshes to videos. Arnab \etal \cite{Arnab_2019_CVPR} presented a two-step approach that involved generating 2D keypoints and initial mesh fits using existing methods, and then using these initial estimates to further refine the results using temporal consistency constraints, e.g., temporal smoothness and 3D priors. However, such a two-step approach is susceptible to errors in either steps and our proposed LMR overcomes this issue with an end-to-end trainable method that provides deeper integration of the temporal data dimension both in training and inference. On the other hand, Kanazawa \etal \cite{humanMotionKanazawa19} and Kocabas \etal \cite{kocabas2019vibe} also presented end-to-end variants of the \textsl{feature-temporal-regressor} where frame-level feature vectors are first encoded using a temporal encoder (e.g., a single recurrent network) and finally processed by a parameter regressor to generate meshes. However, such a global approach to modeling motion dynamics (with only one RNN) does not capture the disparities in locally varying dynamics (e.g., hands vs. legs) which is typically the case in natural human motion. LMR addresses this issue by design with multiple local RNNs in its architecture, one for each pre-defined part of the human body. Such a design also makes mesh parameter regression more amenable by grounding this task in the geometry of the problem, i.e., the SMPL model itself.

\section{Technical Approach}

\begin{figure*}[h!]
\centering
\includegraphics[width=1.0\textwidth]{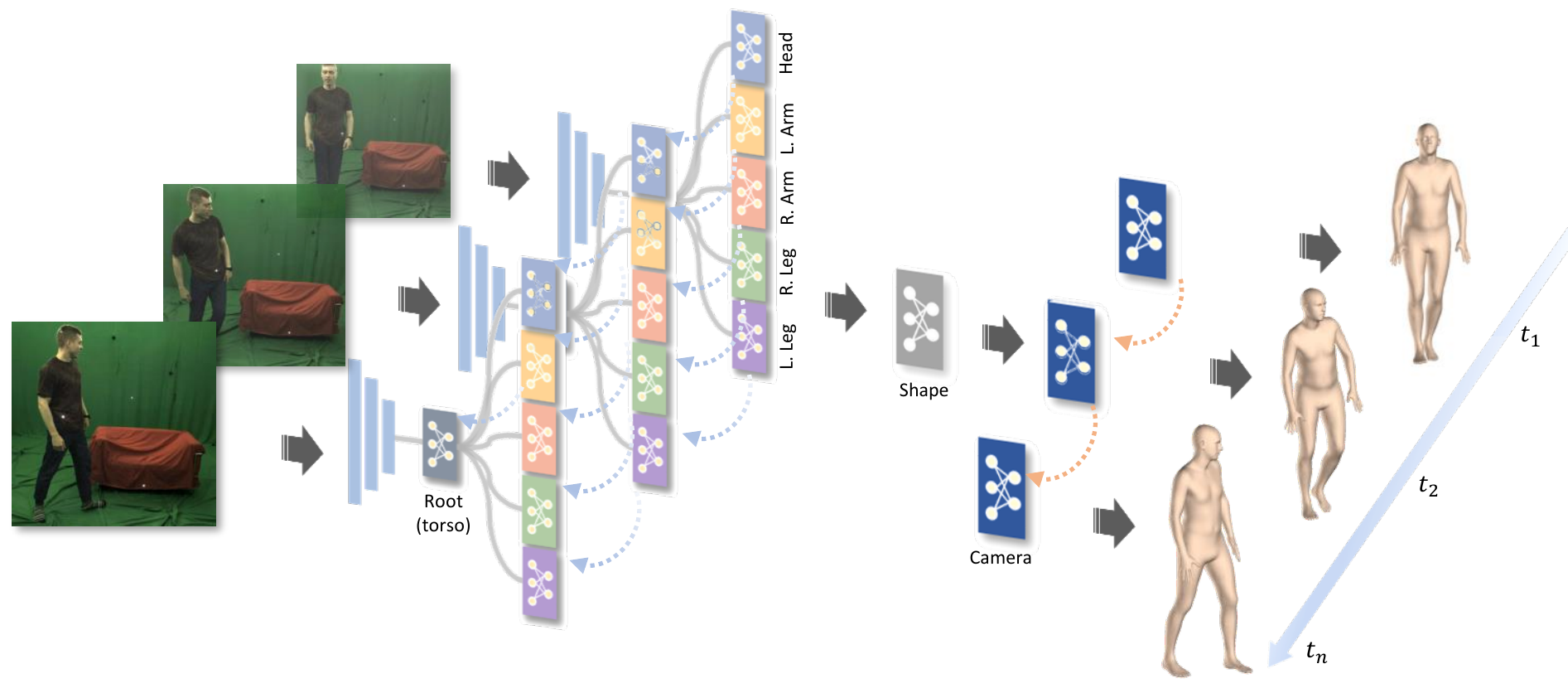}
\caption{The proposed local recurrent modeling approach to human mesh recovery.}
\label{fig:pipeline}
\end{figure*}

\subsection{Parametric Mesh Representation}
\label{sec:smpl}
We use the Skinned Multi-Person Linear (SMPL) model \cite{loper2015smpl} to parameterize the human body. SMPL uses two sets of parameter vectors to capture variations in the human body: shape and pose. The shape of the human body is represented using a 10-dimensional vector $\bm{\beta} \in \mathbb{R}^{10}$ whereas the pose of the body is represented using a 72-dimensional vector $\bm{\theta} \in \mathbb{R}^{72}$. While $\bm{\beta}$ corresponds to the first ten dimensions of the PCA projection of a shape space, $\bm{\theta}$ captures, in axis-angle format \cite{caccavale1999six}, the global rotation of the root joint (3 values) and relative (to the root) rotations of 23 other body joints (69 values). Given $\bm{\beta}$, $\bm{\theta}$, and a learned model parameter set $\bm{\psi}$, SMPL defines the mapping $M(\bm{\beta},\bm{\theta},\bm{\psi}): \mathbb{R}^{82}\rightarrow \mathbb{R}^{3\times N}$ from the 82-dimensional parametric space to a vertex space of $N=6890$ 3D mesh vertices. One can then infer the 24 3D joints of interest (e.g., hips, legs, etc.) $\bm{X} \in \mathbb{R}^{3\times K}, K=24$ using a pre-learned joint regression matrix $\bm{W}$ as $\bm{X}=\bm{W}\bm{J}$. Using a known camera model, e.g., a weak-perspective model as in prior work \cite{hmrKanazawa17}, one can then obtain the corresponding 24 2D image points $\bm{x} \in \mathbb{R}^{2\times K}$ as:
\begin{equation}\label{eq:projection}
    \bm{x}=s\Pi(\bm{X}(\bm{\beta},\bm{\theta})) + \bm{t},
\end{equation}
where the scale $s\in \mathbb{R}$ and translation $\bm{t}\in \mathbb{R}^2$ represent the camera model, and $\Pi$ is an orthographic projection. Therefore, fitting 3D SMPL mesh to a single image involves estimating the parameter set $\bm{\Theta}=\{\bm{\beta},\bm{\theta},s,\bm{t}\}$. In video mesh recovery, we take this a step forward by estimating $\bm{\Theta}$ for every frame in the video. 

\subsection{Learning Local Recurrent Models}

As noted in Section~\ref{sec:intro}, existing video mesh fitting methods formulate the problem in the \textsl{feature-temporal-regressor} design where all motion dynamics in the video are captured using a single RNN. We argue that this is insufficient for mesh estimation due to the inherently complex nature of human actions/motion, more so in challenging in-the-wild scenarios. Our key insight is that natural human motion dynamics has a more locally varying characteristic that can more precisely be captured using locally learned recurrent networks. We then translate this idea into a conditional local recurrent architecture, called \textbf{LMR} and visually summarized in Figure~\ref{fig:pipeline}, where we define multiple recurrent models, one each to capture the dynamics of the corresponding local region in the human body. During training and inference, LMR takes as input a segment of an input video $\bm{V}=\{\bm{I}_1,\bm{I}_2,\ldots,\bm{I}_t, t=1,2,\ldots,T\}$, where $T$ is a design parameter corresponding to the length of the input sequence. LMR first processes each frame with its feature extraction module to produce frame-level feature vectors $\bm{\Phi}=\{\bm{\phi}_1,\bm{\phi}_2,\dots,\bm{\phi}_t\}$ for each of the $T$ frames. LMR then processes $\bm{\Phi}$ with its local part-level recurrent models and associated parameter regressors, and aggregates all part-level outputs to obtain the mesh and camera parameters $\bm{\Theta}_{t}, t=1,2,\ldots,T$ for each frame, finally producing the output video mesh.

\subsubsection{LMR Architecture}
As shown in Figure~\ref{fig:pipeline}(a), our architecture comprises a feature extractor followed by our proposed LMR module. The LMR module is responsible for processing the frame-level representation $\bm{\Phi}$ to output the per-frame parameter vectors $\bm{\Theta}_{t}$. Following the design of the SMPL model and prior work \cite{loper2015smpl,georgakis2020hierarchical}, we divide the human body into six local parts- \textit{root} (4 joints in the root region), \textit{head} (2 joints in the head region), \textit{left arm} (5 joints on left arm), \textit{right arm} (5 joints on right arm), \textit{left leg} (4 joints on left leg), and \textit{right leg} (4 joints on right leg). Given this division, the pose of local part $p_{i}, i=1, \ldots, 6$ can be expressed as $\bm{\theta}^{i}=[\bm{r}_1,\ldots,\bm{r}_{n_{i}}], i=1, \ldots, 6$, where $\bm{r}_{q}$ ($q=1,\ldots,n_{i}$) is a rotation parameterization (e.g., $\bm{r}_{q} \in \mathbb{R}^{3}$ in case of axis angle) of joint $q$ and $n_i$ is the number of joints defined in part $i$. The overall pose parameter vector $\bm{\theta}$ can then be aggregated as $\bm{\theta}=[\bm{\theta}^{1},\ldots,\bm{\theta}^{6}]$.

To capture locally varying dynamics across the video sequence, LMR defines one recurrent model for each of the six parts defined above (see Figure~\ref{fig:pipeline}(b)). The recurrent model for part $i$ is responsible for predicting its corresponding $\bm{\theta}^{i}$. To capture the conditional dependence between parts, the information propagation during training and inference is defined as follows. Given the frame-level feature representation $\bm{\Phi}$, the mean pose vector $\bm{\theta}_{\text{mean}}$, and the mean shape vector $\bm{\beta}_{\text{mean}}$ (note that it is common \cite{hmrKanazawa17,humanMotionKanazawa19,kocabas2019vibe} to initialize mesh fitting with these mean values), the recurrent model responsible for the root part (number 1) first predicts its corresponding pose vector $\bm{\theta}^{1}_{t}, t=1,\ldots,T$ for each of the $t$ frames using the concatenated vector $[\bm{\Phi}_{t},\bm{\theta}^{1}_{\text{mean}},\bm{\beta}_{\text{mean}}]$ as input for the current frame $t$. Note that $\bm{\Phi}_{t}$ is the feature vector for frame $t$ and $\bm{\theta}^{1}_{\text{mean}}$ represents the mean pose parameters of part $p_{1}$. All other recurrent models (parts 2 through 6) then take in as input the concatenated vector $[\bm{\Phi}_{t},\bm{\theta}^{k}_{\text{mean}},\bm{\beta}_{\text{mean}},\bm{\theta}^{1}_t]$ in predicting their corresponding pose vectors $\bm{\theta}^{k}_{t}, k=2,\ldots,6$ and $t=1, \ldots, T$, where $\bm{\theta}^{k}_{\text{mean}}$ represents the mean pose parameters of part $p_{k}$. Note this explicit dependence of part $k$ on the root (part $1$) prediction $\bm{\theta}^{1}$. Given the aggregated (over all 6 parts) pose vector $\bm{\theta}_{t}$, LMR has a fully-connected module that takes as input the concatenated vector $[\bm{\Phi}_{t},\bm{\theta}_{t},\bm{\beta}_{\text{mean}}]$ for each frame $t$ to predict the per-frame shape vectors $\bm{\beta}_{t}, t=1, \ldots, T$. Finally, given an initialization for the camera model $\bm{c}_{\text{init}}=[s_{\text{init}},\bm{t}_{\text{init}}]$, LMR uses the concatenated vector $[\bm{\Phi}_{t},\bm{\theta}_{t},\bm{\beta}_{t},\bm{c}_{\text{init}}]$ as part of its camera recurrent model to predict the camera model $\bm{c}_{t}, t=1, \ldots, T$ for each frame. Note that while we have simplified the discussion and notation here for clarity of exposition, LMR actually processes each batch of input in an iterative fashion, which we next describe in more mathematical detail.

\subsubsection{Training an LMR model} 
As noted above and in Figure~\ref{fig:pipeline}, the proposed LMR module takes as input the video feature set $\bm{\Phi}$ and the mean pose and shape parameters $\bm{\theta}_{mean}$ and $\bm{\beta}_{mean}$ and produces the set of parameter vectors $\bm{\Theta}_{t}=[\bm{\theta}_{t},\bm{\beta}_{t},\bm{c}_{t}]$ for each frame $t$. The LMR block processes each input set in an iterative fashion, with the output after each iteration being used as a new initialization point to further refine the result. The final output $\bm{\Theta}_{t}$ is then obtained at the end of $L$ such iterations. Here, we provide further details of this training strategy. 

Let each iteration step above be denoted by the letter $v$. At step $v=0$, the initial pose and shape values for frame $t$ will then be $\bm{\theta}_{t,v}=\bm{\theta}_\text{mean}$ and $\bm{\beta}_{t,v}=\bm{\beta}_\text{mean}$. The $t,v$ notation refers to the $v^{th}$ iterative step of LMR for frame number $t$. So, given $\bm{\Phi}$, $\bm{\beta}_{t,v}$, and the root pose $\bm{\theta}^{1}_{t,v}$ (recall root is part number 1 from above), the input to the root RNN will be the set of $t$ vectors $[\bm{\Phi}_{t},\bm{\theta}^{1}_{t,v},\bm{\beta}_{t,v}]$ for each of the $t$ frames. The root RNN then estimates an intermediate residual pose $\Delta\bm{\theta}^{1}_{t,v}$, which is added to the input $\bm{\theta}^{1}_{t,v}$ to give the root RNN output $\bm{\theta}^{1}_{t,v}=\bm{\theta}^{1}_{t,v}+\Delta\bm{\theta}^{1}_{t,v}$.

Given the root prediction $\bm{\theta}^{1}_{t,v}$ at iteration $v$, each of the other dependent part RNNs then use this information to produce their corresponding pose outputs. Specifically, for part RNN $k$, the input vector set (across the $t$ frames) will be $[\bm{\Phi}_{t},\bm{\theta}^{k}_{t,v},\bm{\beta}_{t,v},\bm{\theta}^{1}_{t,v}]$ for $k=2,\ldots,6$. Each part RNN first gives its corresponding intermediate residual pose $\Delta\bm{\theta}^{k}_{t,v}$. This is then added to its corresponding input part pose, giving the outputs $\bm{\theta}^{k}_{t,v}=\bm{\theta}^{k}_{t,v}+\Delta\bm{\theta}^{k}_{t,v}$ for $k=2,\ldots,6$. 

After producing all the updated pose values at iteration $v=0$, LMR then updates the shape values. Recall that the shape initialization used at $v=0$ is $\bm{\beta}_{t,v}=\bm{\beta}_\text{mean}$. Given $\bm{\Phi}$, the updated and aggregated pose vector set $\bm{\theta}_{t,v}=[\bm{\theta}^{1}_{t,v},\ldots,\bm{\theta}^{6}_{t,v}]$, and the shape vector set $\bm{\beta}_\text{mean}$, LMR then uses the input vector set $[\bm{\Phi}_{t},\bm{\theta}_{t,v},\bm{\beta}_\text{mean}]$ as part of the shape update module to produce the new shape vector set $\bm{\beta}_{t,v}$ for each frame $t$ during the iteration $v$. 

Given these updated $\bm{\theta}_{t,v}$ and $\bm{\beta}_{t,v}$, LMR then updates the camera model parameters (used for image projection) with a camera model RNN. We use an RNN to model the camera dynamics to cover scenarios where the camera might be moving, although a non-dynamical fully-connected neural network can also be used in cases where the camera is known to be static. Given an initialization for the camera model $\bm{c}_{t,v}=\bm{c}_\text{init}$ at iteration $v=0$, the camera RNN processes the input vector set $[\bm{\Phi}_{t},\bm{\theta}_{t,v},\bm{\beta}_{t,v},\bm{c}_\text{init}]$ to produce the new camera model set $\bm{c}_{t,v}$ for each frame $t$.

After going through one round of pose update, shape update, and camera update as noted above, LMR then re-initializes this prediction process with the updated pose and shape vectors from the previous iteration. Specifically, given the updated $\bm{\theta}_{t,v}$ and $\bm{\beta}_{t,v}$ at the end of iteration $v=0$, the root RNN at iteration $v=1$ then takes as input the set $[\bm{\Phi}_{t},\bm{\theta}^{1}_{t,v},\bm{\beta}_{t,v}]$, where the pose and shape values are not the mean vectors (as in iteration $v=0$) but the updated vectors from iteration $v=0$. LMR repeats this process for a total of $V$ iterations, finally producing the parameter set $\bm{\Theta}_{t}=[\bm{\theta}_{t},\bm{\beta}_{t},\bm{c}_{t}]$ for each frame $t$. Note that this iterative strategy is similar in spirit to the iterative error feedback strategies commonly used in pose estimators \cite{dollar2010cascaded,oberweger2015training,carreira2016human,hmrKanazawa17}.

All the predictions above are supervised using several cost functions. First, if ground-truth SMPL model parameters $\bm{\Theta}^{gt}_{t}$ are available, we enforce a Euclidean loss between the predicted and the ground-truth set:
\begin{equation}
L_\text{smpl}=\frac{1}{T}\sum_{t=1}^{T}\|\bm{\Theta}^{gt}_{t}-\bm{\Theta}_{t}\|_{2}
\label{eq:smpl}
\end{equation}
where the summation is over the $t=T$ input frames in the current batch of data. 

Next, if ground-truth 3D joints $\bm{X}^{gt}_{t} \in \mathbb{R}^{3\times K}$ (recall K=24 from Section~\ref{sec:smpl}) are available, we enforce a mean per-joint L1 loss between the prediction 3D joints $\bm{X}_{t} \in \mathbb{R}^{3 \times K}$ and $\bm{X}^{gt}_{t}$. To compute $\bm{X}_{t}$, we use the predicted parameter set $\bm{\Theta}_{t}$ and the SMPL vertex mapping function $M(\bm{\beta},\bm{\theta},\bm{\psi}): \mathbb{R}^{82}\rightarrow \mathbb{R}^{3\times N}$ and the joint regression matrix $\bm{W}$ (see Section~\ref{sec:smpl}). The loss then is:

\begin{equation}
L_\text{3D}=\frac{1}{T}\frac{1}{K}\sum_{t=1}^{T}\sum_{k=1}^{K}\|\bm{X}^{gt}_{k,t}-\bm{X}_{k,t}\|_{1}
\label{eq:3d}
\end{equation}
where each column of $\bm{X}^{gt}_{k,t} \in \mathbb{R}^{3}$ and $\bm{X}_{k,t} \in \mathbb{R}^{3}$ is one of $K$ joints in three dimensions and the outer summation is over $t=T$ frames as above.

Finally, to provide supervision for camera prediction, we also enforce a mean per-joint L1 loss between the prediction 2D joints $\bm{x}_{t} \in \mathbb{R}^{2 \times K}$ and the ground-truth 2D joints $\bm{x}^{gt}_{t}$. To compute $\bm{x}_{t}$, we use the 3D joints prediction $\bm{X}_{t}$ and the camera prediction $\bm{c}_{t}$ to perform an orthographic projection following Equation~\ref{eq:projection}. The loss then is:

\begin{equation}
L_\text{2D}=\frac{1}{T}\frac{1}{K}\sum_{t=1}^{T}\sum_{k=1}^{K}\|\bm{x}^{gt}_{k,t}-\bm{x}_{k,t}\|_{1}
\label{eq:2d}
\end{equation}
where each column $\bm{x}^{gt}_{k,t} \in \mathbb{R}^{2}$ and $\bm{x}_{k,t} \in \mathbb{R}^{2}$ of $\bm{x}^{gt}_{t}$ and $\bm{x}_{t}$ respectively is one of $K$ joints on the image and the outer summation is over $t=T$ frames as above.

The overall LMR training objective then is:

\begin{equation}
L_\text{LMR}=w_\text{smpl}L_\text{smpl}+w_\text{3D}L_\text{3D}+w_\text{2D}L_\text{2D}
\label{eq:lmr}
\end{equation}
where $w_\text{smpl}$, $w_\text{3D}$, and $w_\text{2D}$ are the corresponding loss weights.

\begin{figure*}[h!]
\centering
\includegraphics[scale=0.532]{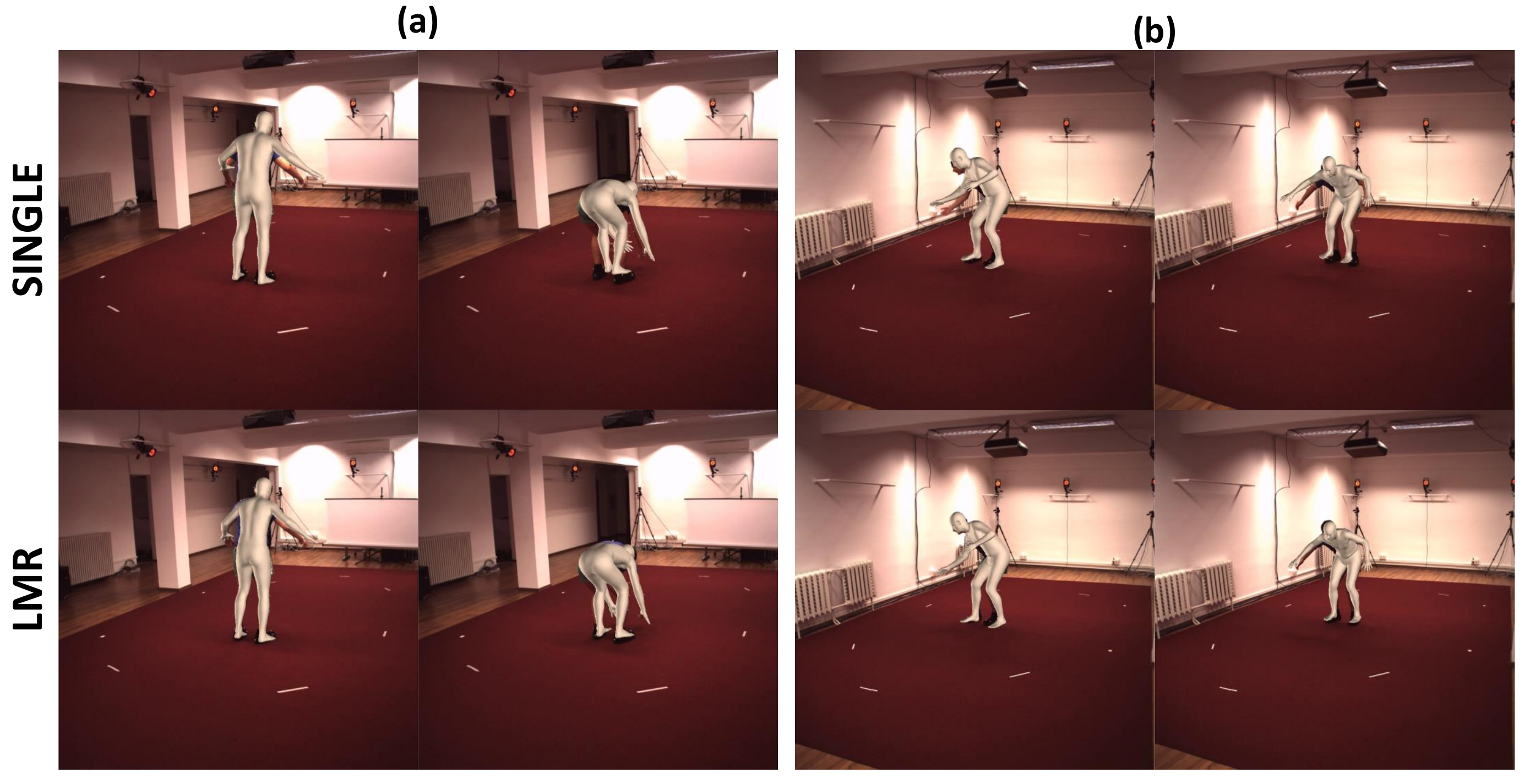}
\caption{Two sets of qualitative results comparing LMR with a single-RNN baseline model.}
\label{fig:qua_w_ablation}
\end{figure*}

\begin{table*}[h!]
    \centering
    \resizebox{0.9\textwidth}{!}{
    \begin{tabular}{c||c|c||c|c||c|c|c|c}
    \hline
    \multirow{2}{*}{Methods} & \multicolumn{2}{c||}{Human3.6M} & \multicolumn{2}{c||}{MPI-INF-3DHP} & \multicolumn{4}{c}{3DPW} \\
    \cline{2-9}
    ~ & MPJPE$\downarrow$ & Rec. Error$\downarrow$ & MPJPE$\downarrow$ & Rec. Error$\downarrow$ & MPJPE$\downarrow$ & Rec. Error$\downarrow$ & PVE$\downarrow$ & Accel$\downarrow$ \\
    \hline
    Single RNN & 69.2 & 45.6 & 100.0 & 66.7 & 87.7 & 55.3 & 101.0 & 19.0 \\
    \hline
    LMR no root dependencies & 66.7 & 43.5 & 97.1 & 64 & 86.3 & 55.1 & 98.9 & 17.6 \\
    \hline
    \textbf{LMR} & \textbf{61.9} & \textbf{42.5} & \textbf{94.6} & \textbf{62.4} & \textbf{81.7} & \textbf{51.2} & \textbf{93.6} & \textbf{15.6} \\
    \hline
    \end{tabular}
    }
    \caption{Results of an ablation study comparing LMR with a single RNN baseline.
    }
    \label{tab:abl_1_quan}
\end{table*}

\begin{figure*}[h!]
\centering
\includegraphics[scale=0.51]{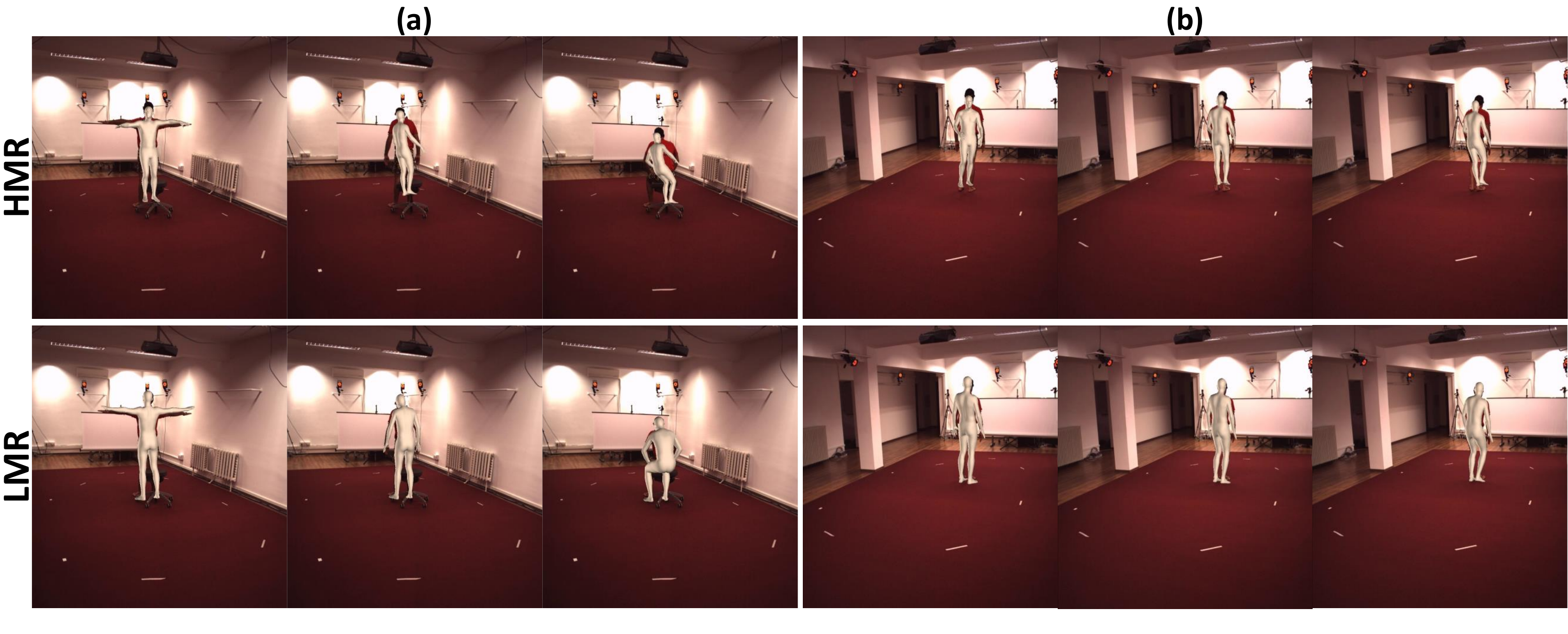}
\caption{Two sets of qualitative results comparing the performance of LMR with the image-based HMR \cite{hmrKanazawa17} method.}
\label{fig:qua_w_hmr}
\end{figure*}

\begin{figure*}[h!]
\centering
\includegraphics[scale=0.51]{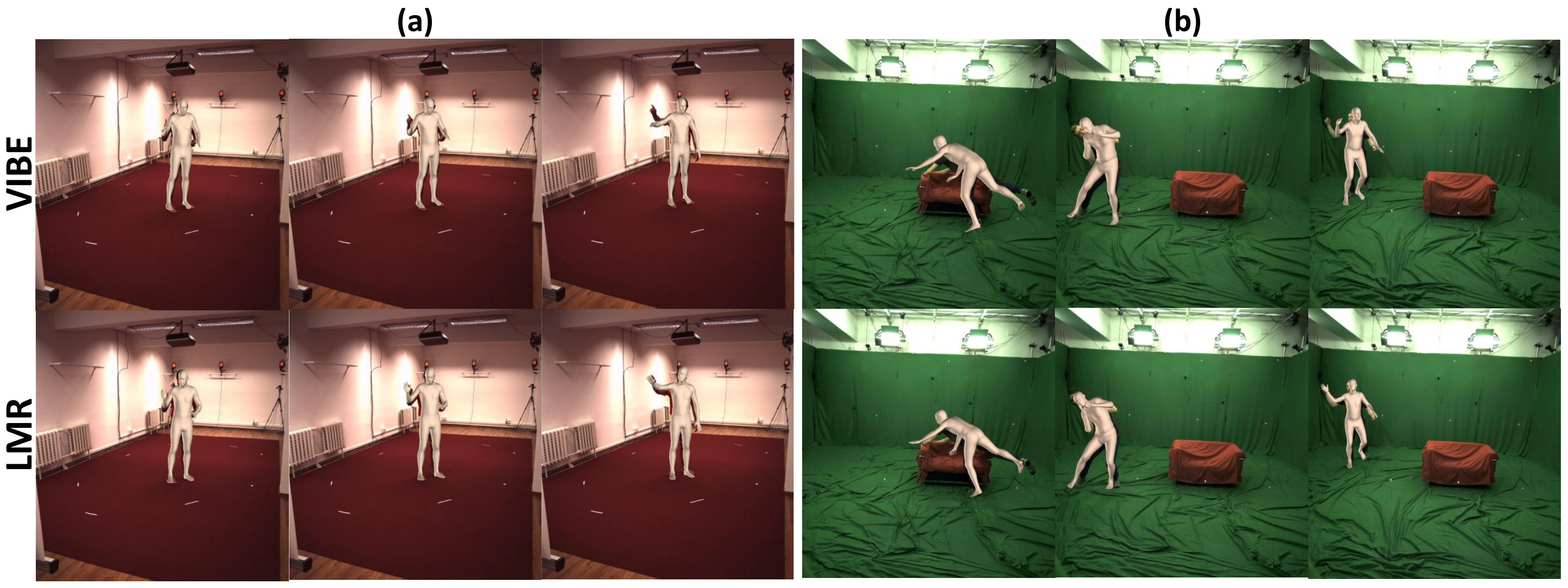}
\caption{Two sets of qualitative results comparing the performance of LMR with the video-based VIBE \cite{kocabas2019vibe} method.}
\label{fig:qua_w_vibe}
\end{figure*}

\begin{table*}[ht]
    \centering
    \resizebox{0.97\textwidth}{!}{
    \begin{tabular}{cc||c|c||c|c||c|c|c|c}
    \hline
    \multicolumn{2}{c||}{\multirow{2}{*}{Methods}} & \multicolumn{2}{|c||}{Human3.6M} & \multicolumn{2}{|c||}{MPI-INF-3DHP} & \multicolumn{4}{c}{3DPW} \\
    \cline{3-10}
    & & MPJPE $\downarrow$ & Rec. Error $\downarrow$ & MPJPE $\downarrow$ & Rec. Error $\downarrow$ & MPJPE $\downarrow$ & Rec. Error $\downarrow$ & PVE $\downarrow$ & Accel $\downarrow$ \\
    \hline
    \multirow{5}{*}{\rotatebox{90}{Image-based}} & Kanazawa \etal \cite{hmrKanazawa17} & 88.0 & 56.8 & 124.2 & 89.8 & 130 & 76.7 & - & 37.4 \\
    \cline{2-10}
    & Omran \etal \cite{omran2018nbf} & - & 59.9 & - & - & - & - & - & - \\
    \cline{2-10}
    & Pavlakos \etal \cite{pavlakos2018learning} & - & 75.9 & - & - & - & - & - & - \\
    \cline{2-10}
    & Kolotouros \etal \cite{kolotouros2019convolutional} & - & 50.1 & - & - & - & 70.2 & - & - \\
    \cline{2-10}
    & Georgakis \etal \cite{georgakis2020hierarchical} & 67.7 & 50.1 & - & - & - & - & - & -\\
    \cline{2-10}
    \hline
    \hline
    Extra-fitting & Kolotouros et al. \cite{kolotouros2019spin} & 62.2 & \textbf{41.1} & 105.2 & 67.5  & 96.9 & 59.2 & 116.4 & 29.8\\
    \hline
    \hline
    \multirow{5}{*}{\rotatebox{90}{Video-based}} & Kanazawa \etal \cite{humanMotionKanazawa19} & - & 56.9 & - & - & 116.5 & 72.6 & 139.3 & \textbf{15.2} \\
    \cline{2-10}
    & Arnab \etal \cite{Arnab_2019_CVPR} & 77.8 & 54.3 & - & - & - & 72.2 & - & - \\
    \cline{2-10}
    & Doersch \etal \cite{NEURIPS2019_d4a93297} & - & - & - & - & - & 74.7 & - & - \\
    \cline{2-10}
    \cline{2-10}
    \cline{2-10}
    \cline{2-10}
    & Kocabas \etal \cite{kocabas2019vibe} & 65.6 & 41.4 & 96.6 & 64.6 & 82.9 & 51.9 & 99.1 & 23.4 \\
    \cline{2-10}
    & \textbf{LMR} & \textbf{61.9} & 42.5 & \textbf{94.6} & \textbf{62.4} & \textbf{81.7} & \textbf{51.2} & \textbf{93.6} & 15.6 \\
    \hline
    \hline
    \end{tabular}
    }
    \caption{Comparing LMR to the state of the art (``-": unavailable result in the corresponding paper).} 
    \label{table:quan}
\end{table*}

\section{Experiments and Results}

\subsection{Datasets and Evaluation} 
Following Kocabas \etal \cite{kocabas2019vibe}, we use a mixture of both datasets with both 2D (e.g., keypoints) as well as 3D (e.g., mesh parameters) annotations. For 2D datasets, we use PennAction \cite{Zhang_2013_ICCV}, PoseTrack \cite{PoseTrack}, and InstaVariety \cite{humanMotionKanazawa19}, whereas for 3D datasets, we use Human3.6M \cite{ionescu2013human3}, MPI-INF-3DHP \cite{mehta2017monocular}, and 3DPW \cite{von2018recovering}. In all our experiments, we use exactly the same settings as Kocabas \etal \cite{kocabas2019vibe} for a fair benchmarking of the results. To report quantitative performance, we use evaluation metrics that are now standard in the human mesh research community. On all the test datasets, we report both mean-per-joint position error (MPJPE) as well as Procrustes-aligned mean-per-joint position error (PA-MPJPE). Additionally, following Kanazawa \etal \cite{humanMotionKanazawa19} and Kocabas \etal \cite{kocabas2019vibe}, on the 3DPW test set, we also report the acceleration error (``Accel."), which is the average (across all keypoints) difference between the ground truth and predicted acceleration of keypoints, and the per-vertex error (PVE). 

\subsection{Ablation Results}
We first present results of an ablation experiment conducted to study the efficacy of the proposed design of LMR, i.e., the use of multiple local recurrent models as opposed to a single recurrent model as is done in prior work \cite{kocabas2019vibe}. Here, we follow the same pipeline as Figure~\ref{fig:pipeline} in spirit, with the only difference being the use of only one RNN to infer all the pose parameters $\bm{\theta}$ instead of the six RNNs depicted in Figure~\ref{fig:pipeline}(b). All other design choices, e.g., for the shape model or the camera model, remain the same as LMR. We show qualitative results of this experiment in Figure~\ref{fig:qua_w_ablation} and quantitative results in Table~\ref{tab:abl_1_quan}. In Figure~\ref{fig:qua_w_ablation}, we show two frames from two different video sequences in (a) and (b). The first row shows results with this single RNN baseline and the second row shows corresponding results with our full model, i.e., LMR. One can note that LMR results in better mesh fits, with more accurate $\bm{\Theta}$-inference in regions such as hands and legs. We further substantiate this performance gap quantitatively in Table~\ref{tab:abl_1_quan}, where one can note the proposed LMR gives consistently better performance than its baseline single RNN counterpart across all datasets as well as evaluation metrics. 

\subsection{Comparison with the state-of-the-art results} 
We compare the performance of LMR with a wide variety of state-of-the-art image-based and video-based methods. We first begin with a discussion on relative qualitative performance. In Figure~\ref{fig:qua_w_hmr}, we show three frames from two different video sequences in (a) and (b) comparing the performance of the image-based HMR method \cite{hmrKanazawa17} (first row) and our proposed LMR. Since LMR is a video-based method, one would expect substantially better performance, including in cases where there are self-occlusions. From Figure~\ref{fig:qua_w_hmr}, one can note this is indeed the case. In the first column of Figure~\ref{fig:qua_w_hmr}, HMR is unable to infer the correct head pose (it infers front facing when the person is actually back back facing), whereas LMR is able to use the video information from prior to this frame to infer the head pose correctly. Note also HMR's incorrect inference in other local regions, e.g., legs, in the subsequent frames in Figure~\ref{fig:qua_w_hmr}(a). This aspect of self-occlusions (i.e., invisible face keypoints) is further demonstrated in Figure~\ref{fig:qua_w_hmr}(b), where HMR is unstable (front facing on a few and back facing on a few frames), whereas LMR consistently infers the correct pose. 

Next, we compare the performance of LMR with the state-of-the-art video-based VIBE method \cite{kocabas2019vibe}. In Figure~\ref{fig:qua_w_vibe}, we show three frames from two different video sequences in (a) and (b). One can note substantial performance improvement in several local regions from these results. In particular, LMR infers more accurate hand pose and camera model parameters in Figure~\ref{fig:qua_w_vibe}(a) when compared to VIBE. The results in Figure~\ref{fig:qua_w_vibe}(b), a more challenging scenario, best illustrates the benefits offered by proposed local design of LMR. Given the variety of body movements in this set of frames, one can note the improved performance of LMR in several regions- hands and legs in the first column, head in the second column, and hands and legs again in the third column. These results are further substantiated in the quantitative comparison we discuss next. 

We provide a quantitative comparison of the performance of LMR to various state-of-the-art image- and video-based methods in Table~\ref{table:quan}. We make several observations. First, as expected, LMR gives substantially better performance when compared to the image-based method of Kanazawa \etal \cite{hmrKanazawa17} (MPJPE of 61.9 mm for LMR vs. 88.0 mm for HMR on Human3.6M, 94.6 mm for LMR vs. 124.2 mm for HMR on MPI-INF-3DHP, and 81.7 mm for LMR vs. 130.0 mm for HMR on 3DPW). This holds with other image-based methods as well (first half of Table~\ref{table:quan}). Next, LMR gives competitive performance when compared to state-of-the-art video-based methods as well. In particular, further substantiating the discussion above, LMR generally outperforms Kocabas \etal \cite{kocabas2019vibe} with margins that are higher on the ``in-the-wild" datasets (MPJPE of 94.6 mm for LMR vs. 96.6 mm for Kocabas \etal \cite{kocabas2019vibe} on MPI-INF-3DHP, Accel. of 15.6 mm/s$^2$ for LMR vs. 23.4 mm/s$^2$ for Kocabas \etal \cite{kocabas2019vibe} on 3DPW), further highlighting the efficacy of LMR's local dynamic modeling. 

Finally, in Table~\ref{table:quan}, we also compare our results with those of Kolotouros \etal \cite{kolotouros2019spin} that uses an additional step of in-the-loop model fitting. Note that despite our proposed LMR \textbf{not} doing this extra model fitting, it outperforms Kolotouros \etal \cite{kolotouros2019spin} in most cases, with particularly substantial performance improvements on MPI-INF-3DHP (MPJPE of 94.6 mm for LMR vs. 105.2 mm for Kolotouros \etal \cite{kolotouros2019spin}) and 3DPW (MPJPE of 81.7 mm for LMR vs. 96.9 mm for Kolotouros \etal \cite{kolotouros2019spin}).

\section{Conclusions}
We considered the problem of video human mesh recovery and noted that the currently dominant design paradigm of using a single dynamical system to model all motion dynamics, in conjunction with a ``flat" parameter regressor is insufficient to tackle challenging in-the-wild scenarios. We presented an alternative design based on local recurrent modeling, resulting in a structure-informed learning architecture where the output of each local recurrent model (representing the corresponding body part) is appropriately conditioned based on the known human kinematic structure. We presented results of an extensive set of experiments on various challenging benchmark datasets to demonstrate the efficacy of the proposed local recurrent modeling approach to video human mesh recovery.

{\small
\bibliographystyle{ieee_fullname}
\bibliography{egbib}
}

\end{document}